\documentclass[journal]{IEEEtran}

\ifCLASSINFOpdf
  \usepackage[pdftex]{graphicx}
\else
\fi

\usepackage[cmex10]{amsmath}
\usepackage{amssymb}
\usepackage{algorithmic}
\usepackage{url}
\usepackage{multirow, subfigure}
\usepackage[procnumbered,ruled]{algorithm2e}

\def\etal{\emph{et al.}~}
\DeclareMathOperator*{\argmax}{arg\,max}
\DeclareMathOperator*{\argmin}{arg\,min}

\usepackage{mathtools}

\DeclarePairedDelimiter\floor{\lfloor}{\rfloor}

\newcommand{\mli}[1]{\mathit{#1}}

\begin{document}

\title{Recognizing and Presenting the Storytelling Video Structure with Deep Multimodal Networks}

\author{Lorenzo~Baraldi,~
        Costantino~Grana,~\IEEEmembership{Member,~IEEE,}
        and~Rita~Cucchiara,~\IEEEmembership{Member,~IEEE}
\thanks{L.~Baraldi, C.~Grana and R.~Cucchiara are with the Department
of Engineering ``Enzo Ferrari'', University of Modena and Reggio Emilia, Italy (e-mail: lorenzo.baraldi@unimore.it; costantino.grana@unimore.it; rita.cucchiara@unimore.it).}
}

\markboth{IEEE Transactions on Multimedia,~Vol.~13, No.~9, September~2014}%
{Baraldi \MakeLowercase{\textit{et al.}}: Bare Demo of IEEEtran.cls for Journals}

\maketitle

\begin{abstract}
This paper presents a novel approach for temporal and semantic segmentation of edited videos into meaningful segments, from the point of view of the storytelling structure. The objective is to decompose a long video into more manageable sequences, which can in turn be used to retrieve the most significant parts of it given a textual query and to provide an effective summarization.
Previous video decomposition methods mainly employed perceptual cues, tackling the problem either as a story change detection, or as a similarity grouping task, and the lack of semantics limited their ability to identify story boundaries. Our proposal connects together perceptual, audio and semantic cues in a specialized deep network architecture designed with a combination of CNNs which generate an appropriate embedding, and clusters shots into connected sequences of semantic scenes, i.e. \textit{stories}. A retrieval presentation strategy is also proposed, by selecting the semantically and aesthetically ``most valuable'' thumbnails to present, considering the query in order to improve the storytelling presentation.
Finally, the subjective nature of the task is considered, by conducting experiments with different annotators and by proposing an algorithm to maximize the agreement between automatic results and human annotators.
\end{abstract}


\IEEEpeerreviewmaketitle

\section{Introduction}
\IEEEPARstart{R}{eal-Time} Entertainment is currently the dominant traffic category on the web, and video accounts for most of it. In the first half of 2016, 71\% of downstream bytes during peak period were due to this category, and the top 3 applications were Netflix (35\%), YouTube (18\%) and Amazon Video (4\%)~\cite{sandvine2016}. 

While User Generated Videos are popular, in a recent survey, people of ages 13-34 indicated that the primary type of video content they viewed was TV shows, full-length movies, music videos, sports, and clips of TV shows for a total of 78\% of the respondents, while another 8\% was ``Videos of people playing video games'', and the rest was ``Other user-generated content''~\cite{tivo2015}.

Browsing video content is not as easy as searching other media, e.g. images. The returned result page in most search engines presents videos through their thumbnails, so assessing if the content is indeed pertinent to our query requires further playing, possibly with fast forward and backwards operations. But professionally edited videos have a well defined storytelling structure that we could leverage for improving the user experience. This structure may be usually described by a hierarchical decomposition: at the lowest level we have frames, which are in turn grouped in shots, a sequence of consecutive frames taken by a single camera act~\cite{petersohn10}. Multiple terms appear in literature to describe \emph{temporal video segments} on the level above shots: scene, topic unit, logical unit, logical story unit, story unit, video paragraph, and macro segment. The term \emph{scene} borrows from stage production, focuses on the location of the action, and is mainly used in fictional narrative-driven videos. \emph{Topic unit}, instead, is focused on the subject discussed, and is usually employed for news, documentaries and educational videos. Logical or story unit, \emph{story} in short, are terms which try to comprehend both settings. 

Edited videos tell us a story by means of many multimedia cues: images, audio, speech and text are combined at acquisition time and during the editing process to create a connected storytelling structure. Indeed, while user-generated videos are usually composed of short sequences with one or few shots, broadcast and professional videos contain long sequences of shots, concatenated together to form a collection of stories. 
Thus we use the term \textit{stories} to define a sequence of shots with a coherent semantics and their related annotations. Then, stories which even not temporally adjacent but semantically similar, could be retrieved by concept-based or similarity-based queries.

Recognizing this storytelling structure from an arbitrary edited video is a challenging problem, both due to the diversity of video domains (news, documentaries, movies, sport videos, etc.), and to the intrinsic subjectivity of the task. Nevertheless, the large heritage of broadcast video footage could considerably benefit of an algorithm capable of solving this task, since that would augment the accessibility and retrieval possibilities in large broadcast videos collections, which are still not as easy to access and search as user generated videos.

In this paper, we address the problem of automatically extracting the storytelling structure of an edited video, by grouping shots in stories with a multimodal deep network approach, which employs semantic, visual, textual and audio cues. We show how the hierarchical decomposition can improve retrieval results presentation with semantically and aesthetically effective thumbnails. 

Here we take into account four issues in the whole detection-annotation-retrieval pipeline. The first, most challenging issue comes from the need of integrating diverse multimedia cues: perceptual uniformity, audio consistency, the persistence of the semantic content expressed by the speaker or recognized in the frames need to be integrated in a common framework. The second is the multimodal temporal synchronization: all these elements are not precisely overlapped in time: concepts expressed by speech are not only related to the single shot they appear in, and background music can overlap with stories with different semantic meaning. A third problem is how to present or summarize stories that can be retrieved by detected concepts. Finally, the never-ending debate on measuring the quality of automatic results with respect to the user perceived quality is addressed.

The main novelties of our work are:
\begin{itemize}
	\item We propose a strategy for extracting semantic features from the video transcript 
	which are incorporated with perceptual cues into a multimodal embedding space, thanks 
	to a Triplet Deep Network. Using these features, we are able to provide a state of the 
	art story detection algorithm.
	\item We leverage the extracted storytelling structure to provide improved query 
	dependent thumbnails, combining semantic and aesthetic information.
	\item We discuss the problem of evaluating story detection and provide a dynamic 
	programming algorithm for managing the subjectivity in presence of different contradicting
	annotations.	
\end{itemize}


\section{Related Work}
\label{sec:related}
In this section we review the literature related to story detection, video retrieval and thumbnail selection techniques.

Existing works in the field of automatic story detection can be roughly categorized into three groups~\cite{hanjalic99}: \textit{rule-based methods}, that consider the way a video is structured in professional movie production, \textit{graph-based methods}, where shots are arranged in a graph representation, and \textit{clustering-based methods}.

The drawback of rule-based methods is that they tend to fail in videos where film-editing rules are not followed strictly, or when two adjacent stories are similar and follow the same rules. 
The method proposed by Liu~\etal in \cite{liu13} falls in this category: they propose a visual based probabilistic framework that imitates the authoring process. In \cite{chasanis09}, shots are represented by means of key-frames, clustered using spectral clustering and low level color features, and then labeled according to the clusters they belong to. Since video editing tends to follow repetitive patterns, boundaries are detected from the alignment score of the symbolic sequences, using the Needleman-Wunsch algorithm. 

In graph-based methods, instead, shots are arranged in a graph representation and then clustered by partitioning the graph. The Shot Transition Graph (STG)~\cite{yeung95} is one of the most used models in this category: here each node represents a shot and the edges between the shots are weighted by shot similarity. In~\cite{rasheed05}, color and motion features are used to represent shot similarity, and the STG is then split into subgraphs by applying the normalized cuts for graph partitioning. Sidiropoulos~\etal\cite{sidiropoulos11} introduced a STG approximation that exploits features from the visual and the auditory channel.

Clustering-based solutions assume that similarity of shots can be used to group them into meaningful clusters, thus directly providing the final temporal boundaries. In \cite{acm}, for instance, a Siamese Network is used together with features extracted from a CNN and time features to learn distances between shots. Spectral clustering is then applied to detect coherent sequences.

Our work belongs to this latter class, but overcomes the limitations of the previous
approaches incorporating audio and video in two flavors: they are used to extract 
perceptual features (e.g. CNN activations and MFCC) and semantic features 
(e.g. concepts and transcript words). We employ a temporal aware clustering algorithm 
which, by construction, generates contiguous segments: temporal coherence is not an
additional requirement forced later, but is optimized during the clustering itself.

\begin{figure*}[t]
\centering
\includegraphics[width=\textwidth]{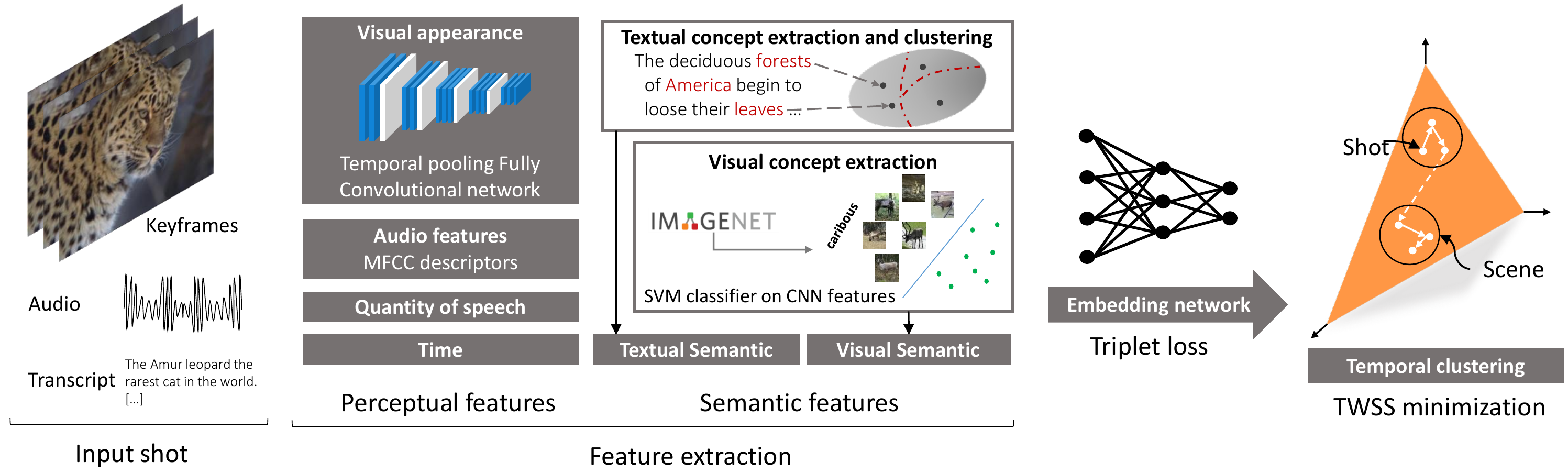}
\caption{Overview of the proposed approach. A semantic embedding is learned through a triplet deep network, which combines perceptual and semantic features. }
\label{fig:schema}
\end{figure*}

Lot of work has also been proposed for video retrieval: with the explosive growth of online videos, this has become a hot topic in computer vision. In their seminal work, Sivic \etal proposed Video Google~\cite{sivic2003video}, a system that retrieves videos from a database via bag-of-words matching. Lew \etal~\cite{lew2006content} reviewed earlier efforts in video retrieval, which mostly relied on feature-based relevance feedback or similar methods.

More recently, concept-based methods have emerged as a popular approach to video retrieval. Snoek \etal~\cite{snoek2007adding} proposed a method based on a set of concept detectors, with the aim to bridge the semantic gap between visual features and high level concepts.  
In~\cite{ballan2015data}, authors proposed a video retrieval approach based on tag propagation: given an input video with user-defined tags, Flickr, Google Images and Bing are mined to collect images with similar tags: these are used to label each temporal segment of the video, so that the method increases the number of tags originally proposed by the users, and localizes them temporally. In~\cite{Lin2014} the problem of retrieving videos using complex natural language queries is tackled, by first parsing the sentential descriptions into a semantic graph, which is then matched to visual concepts using a generalized bipartite matching algorithm. This also allows to retrieve the relevant video segment given a text query. Our method, in contrast to~\cite{ballan2015data}, does not need any kind of initial manual annotation, and, thanks to the availability of the video structure, is able to return specific stories related to the user query. This provides the retrieved result with a context that allows to better understand the video content.

Retrieved results need eventually to be presented to the user, but previewing many videos playing simultaneously is not something feasible. The usual approach is to present a set of video thumbnails. Thumbnails are basically surrogates for videos~\cite{craggs2014thumbreels}, as they take the place of a video in search results. Therefore, they may not accurately represent the content of the video, and create an \textit{intention gap}, i.e. a discrepancy between the information sought by the user and the actual content of the video. Most conventional methods aim at selecting the ``best'' thumbnail, and have focused on learning visual representativeness purely from visual content~\cite{kang2005learn,rav2006making}. However, more recent researches have focused on choosing query-dependent thumbnails to supply specific thumbnails for different queries.
To reduce the intention gap, \cite{craggs2014thumbreels} proposes a new kind of animated preview, constructed of frames taken from a full video, and a crowdsourced tagging process which enables the matching between query terms and videos. Their system, while going in the right direction, suffers from the need of manual annotations, which are often expensive and difficult to obtain.

In~\cite{liu2011query}, instead, authors proposed a method to enforce the representativeness of a selected thumbnail given a user query, by using a reinforcement algorithm to rank frames in each video and a relevance model to calculate the similarity between the video frames and the query keywords. Recently, Liu \etal~\cite{liu2015multi} trained a deep visual-semantic embedding to retrieve query-dependent video thumbnails. Their method employs a deeply-learned model to directly compute the similarity between a query and video thumbnails, by mapping them into a common latent semantic space.

Our work can push things further, because we already retrieved a video story for which the query is relevant, thus we just need to pick a keyframe within a very limited set of candidates. All possible thumbnails are thus ranked according to their relevance to the query and to their aesthetic value, providing the best presentation of the result for the specific user request.

\section{Perceptual-Semantic Feature Embedding and Clustering for Story Detection}
\label{sec:scene_detection}

We tackle the task of detecting stories in edited videos as a supervised temporally constrained clustering problem. We firstly extract a rich set of perceptual and semantic features from each shot. 
In order to obtain a significant measure of similarity between shots features, we learn an embedding of these features in a Euclidean space. 
Finally we detect the optimal story boundaries by minimizing the sum of squared distances inside temporal segments (candidate stories), using a penalty term to automatically select the number of stories. 
A summary of our approach is depicted in Fig.~\ref{fig:schema}.

In the following, we present a set of perceptual features based on visual appearance, audio, speech and time. Then, we propose two semantic features which rely on a joint conceptual analysis of the visual content and of the transcript, and which account for story changes which are not recognizable using purely perceptual cues. Eventually, we present the embedding and clustering strategies.


\subsection{Perceptual features}

\subsubsection*{Visual appearance}
A shot in an edited video is usually uniform from the visual content point of view, and it is therefore reasonable to rely on key-frames to describe visual appearance. At the same time, using a single key-frame could result in a poor description of both short and long shots, since the visual quality could be unsatisfactory, or its content may be insufficient to describe the temporal evolution of a shot. For this reason, we propose a solution which preserves the ability of Convolutional Neural Networks (CNNs) to extract high level features, while accounting for the temporal evolution of a shot.

Specifically, we build a Temporal Pooling Fully Convolutional Neural Network, which can encode the visual appearance of a variable number of key-frames into a descriptor with fixed size. The proposed network is Fully Convolutional in that it contains only convolutional and pooling stages, and does not include fully connected layers. Moreover, the last stage of the network performs a temporal pooling operation, thus reducing a variable number of key-frames to a fixed dimension.

The architecture of the network follows that of the 16 layers model from VGG~\cite{simonyan2014very}
To keep a fully convolutional architecture, the last fully connected layers are removed, and a temporal pooling layer is added at the end. Parameters of the network are initialized with those pre-trained on the ILSVRC-12 dataset~\cite{ILSVRC15}.

Given a set of key-frames $\{ I_1, ..., I_t\}$ with size $l \times l$, each of them is independently processed by the convolutional and spatial pooling layers of the network, thus obtaining a three-dimensional tensor $\mli{CNN}(I_i)$ for each key-frame $I_i$ with shape $\floor*{\frac{l}{f}} \times \floor*{\frac{l}{f}} \times k $, where $f$ is the factor by which the input image is resized by the spatial pooling layers of the network, and $k$ is the number of convolutional filters of the last layer. Each of these $k$ activation maps intuitively contains the spatial response of a specific high level feature detector over the input image. The temporal pooling layer performs a max-pooling operation over time: the output of this layer, therefore, has the same shape of $\mli{CNN}(I_i)$, and contains, in each position $(x,y,j)$, the element-wise maximum along the time dimension, $\max_{i \in \{1, ..., t\}} \mli{CNN}(I_i)(x,y,j)$. For the VGG-16 model, the input shape is $224\times 224$, the resize factor $f$ is 32, and $k$ is 512.

Based on a preliminary evaluation, we chose to extract three key-frames per shot, with uniform sampling. More sophisticated sampling techniques were also tested: we encoded all the frames in a shot using color histogram and selected the $t$ most different key-frames. However, no significant improvement with respect to uniform sampling was observed. Average-pooling in the temporal layer was also tested, but it led to worse performance than max-pooling.

\subsubsection*{Audio features}
The audio of an edited video is another meaningful cue for detecting story boundaries, since audio effects and soundtracks are often used in professional video production to underline the development of a story, and a change in soundtrack usually highlights a change of content. For this reason, a standard audio descriptor based on short-term power spectrum is employed.

Following recent works in the field~\cite{habibian2015videostory}, we extract MFCCs descriptors~\cite{logan2000mel} over a 10ms window. The descriptors consist of 13 values, 30 coefficients and the log-energy, along with their derivatives and the
second derivatives. The MFCC descriptors are aggregated by Fisher vectors using a Gaussian Mixture Model with 256 components, resulting in a 46,080 dimensional vector.

\subsubsection*{Quantity of speech}
Sometimes a pause in the speaker discourse can be enough to identify a change of story: for this reason, we turn to the video transcript and build a quantity of speech feature, which computes the amount of words being said inside a shot. Notice that, when the video transcript is not directly provided by the video producer, it can be obtained with standard speech-to-text techniques.

For each shot, the quantity of speech is defined as the number of words which appear in that shot, normalized with respect to the maximum number of words found in a shot for the full video.

\subsubsection*{Time features}
We also include the timestamp and length of each shot. The rationale behind this choice is that since stories need to be temporally consecutive, shots having similar semantic content which are temporally distant should be distinguishable. Moreover, the average length of stories can be a useful prior to be learned.

Notice that a shot-based representation has been kept in all the proposed features. For each shot, indeed, the concatenation of its feature vectors will be the input of the Triplet Deep Network, which will learn the embedding.

\subsection{Semantic features}
Perceptual features can be sufficient to perform story detection on videos which have a simple storyline; however, it is often the case that story boundaries correspond to changes in topic
which can not be detected by simply looking at appearance and sound. In the following we extract concepts from the video transcript, and project them into a semantic space; each concept is then validated by looking at the visual content of a shot.

To collect candidate concepts, sentences in the transcript are firstly parsed and unigrams which are annotated as \textit{noun}, \textit{proper noun} and \textit{foreign word} are collected with the Stanford CoreNLP~\cite{de2006generating} part-of-speech tagger. Selected unigrams contain terms which may be present inside the video, and may be helpful to visually detect a change in topic. On the contrary, there are also terms which do not have concrete visual patterns, but that can still be important to infer a change in topic from the transcript. We will describe two features to account for both these situations.

\subsubsection*{Concept clustering}
The resulting set of terms can be quite redundant and contain lots of synonyms, therefore we cluster it according to the pairwise similarities of terms, in order to obtain a set of semantically non-related clusters. In particular, we train a Word2Vec model~\cite{mikolov2013efficient} on the dump of the English Wikipedia. The basic idea of this model is to fit a word embedding such that the words in corpus can predict their context with high probability. Semantically similar words lie close to each other in the embedded space.

In our case, each word is mapped to a $1000$-dimensional feature vector, and the semantic similarity of two terms is defined as the cosine similarity between their embeddings. The resulting similarity matrix is then used together with spectral clustering to cluster the mined terms into $K$ concept groups. $K$ was set to 50 in all our experiments.

Due to the huge variety of concepts which can be found in the video collection, the video corpus itself may not be sufficient to train detectors for the visual concepts. Therefore, we mine images from the Imagenet database~\cite{deng2009imagenet}, which contains images from more than 40.000 categories from the WordNet~\cite{miller1995wordnet} hierarchy. Our method, in principle, is applicable to any visual corpus, provided that it contains a sufficiently large number of categories.

Each concept in Imagenet is described by a set of words or word phrases (called \textit{synset}). We match each unigram extracted from the text with the most similar synset in the aforementioned semantic space, and call $M(u)$ the synset resulting from this matching process for a unigram $u$. For synsets containing more than one word, we take the average of the vectors from each word and $L_2$-normalize the resulting vector.

\subsubsection*{Visual semantic features}
\label{sec:visual_semantic}
Having mapped each concept from the video transcript to an external corpus, a classifier can be built to detect the presence of a visual concept in a shot. Since the number of terms mined from text data is large, the classifier needs to be efficient. Images from the external corpus are represented using feature activations from pre-trained deep convolutional neural networks. Then, a linear probabilistic SVM is trained for each concept, using randomly sampled negative training data; the probability output of each classifier is then used as an indicator of the presence of a concept in a shot. Again the 16-layers model from VGG~\cite{simonyan2014very} is employed, pretrained on the ILSVRC-2012~\cite{ILSVRC15} dataset. We use the activations from layer \texttt{fc6}.

We build a feature vector which encodes the influence of each concept group on the considered shot. Given the temporal coherency of a video, it is unlikely for a visual concept to appear in a shot which is far from the point in which the concept was found in the transcript. At the same time, concepts expressed in the transcript are not only related to the single shot they appear in, but also to its neighborhood. For this reason, we apply a normalized Gaussian weight to each term based on the temporal distance. Formally, the probability that a term $u$ is present in a shot $s$ is defined as:
\begin{equation}
\label{eq:proba_unigram}
P(s, u) = f_{M(u)}(s) \ e^{-\frac{(t_u - t_s)^2}{2\sigma_a^2}}
\end{equation}
where $M$ is the mapping function to the external corpus, and $f_{M(u)}(s)$ is the probability given by the SVM classifier trained on concept $M(u)$ and tested on shot $s$. $t_u$ and $t_s$ are the timestamps of term $u$ and shot $s$ (expressed as frame indexes). Parameter $\sigma_a$ was set as 20 times the frame rate in all experiments, so to have a full width at half maximum of the Gaussian equal to $2\sqrt{2\ln(2)}\cdot 20 \approx 47$ seconds. 

Given the definition of $P(s, u)$ the visual concept feature of a shot is a $K$-dimensional vector, defined as
\begin{equation}
\mathbf{v}(s) = \left[ \sum_{u \in \mathcal{T}} \delta_{u,i}  P(s, u) \right]_{i=1,...,K}
\end{equation} 
where $\mathcal{T}$ is the set of all terms inside a video, $\delta_{u,i} \in \{ 0,1\}$ indicates whether term $u$ belongs to the $i$-th concept group. 

\subsubsection*{Textual semantic features}
Textual concepts are as important as visual concepts to detect story changes, and detected concept groups provide an ideal mean to describe topic changes in text. Therefore, a textual concept feature vector, $\mathbf{t}(s)$, is built as the textual counterpart of $\mathbf{v}(s)$
\begin{equation}
\mathbf{t}(s) = \left[ \sum_{u \in \mathcal{T}} \delta_{u,i} e^{-\frac{(t_u - t_s)^2}{2\sigma_a^2}} \right]_{i=1,...,K}
\end{equation}
We thus get a representation of how much each concept group is present in the transcript of a shot and in its neighborhood.

The overall feature vector $\mathbf{x}$ of a shot $s$ is the concatenation of all the perceptual and conceptual features.

\subsection{Embedding network}
\label{sec:embedding_network}
Given an input video, we would like to partition it into a set of sequences with the goal of maximizing the semantic coherence of the resulting segments. 
To this end we would need a distance between shots feature vectors $\mathbf{x}$, which reflects the semantic similarity. Instead of explicitly defining this hypothetical distance, we learn an embedding function $\phi(\mathbf{x})$ that maps the feature vector of a shot to a space in which the Euclidean distance has the required semantic properties.

The ideal pairwise distance matrix would be $[ \| \phi(\mathbf{x}_i) - \phi(\mathbf{x}_j) \|^2]_{i,j=1,...,n} = [1-\delta_{i,j}]_{i,j=1,...,n}$, where $\delta_{i,j}$ is a binary function that indicates whether shot $i$ and shot $j$ belong to the same story. 
For this reason, $\phi(\cdot)$ is learned such that a shot $\mathbf{x}_i$ of a specific story should be closer to all the shots $\mathbf{x}_i^+$ of the same story than to any shot $\mathbf{x}_i^-$ of any other story, thus enforcing $\| \phi(\mathbf{x}_i) - \phi(\mathbf{x}_i^+) \|^2 < \| \phi(\mathbf{x}_i) - \phi(\mathbf{x}_i^-) \|^2$.

\begin{algorithm}[tb]
    \SetKwInOut{Input}{Input}
    \SetKwInOut{Output}{Output}
    \Input{Number of iterations $T$; mini-batch size $N$; regularization strength $\lambda$; learning rate $\eta$; momentum $\gamma$; training triplets $( \mathbf{x}_i,\mathbf{x}_i^+,\mathbf{x}_i^-)_{i}$}

    \Output{Optimized parameters $\mathbf{w}$ and $\mathbf{\theta}$}
    
    Initialize $\mathbf{w}$ according to \cite{glorot2010understanding} and $\mathbf{\theta}$ to $\vec{0}$.
    
    \For{$1\leq t \leq T$}{
        Randomly select $N$ training triplets
        
        \For{$1 \leq i \leq N$}{
            \If{$\| \phi(\mathbf{x}_i) - \phi(\mathbf{x}_i^+) \|^2 + \left(1 - \| \left. \phi(\mathbf{x}_i) - \phi(\mathbf{x}_i^- \right) \|^2 \right) > 0$}{
$\mathbf{v}_w \gets \gamma \mathbf{v}_w + \eta \left(\lambda \mathbf{w} + \frac{1}{N}\sum_{i=1}^N \frac{\partial L_i}{\partial \mathbf{w}} \right)$\\
$\mathbf{w} \leftarrow \mathbf{w} - \mathbf{v}_w$\\
$\mathbf{v}_\theta \gets \gamma \mathbf{v}_\theta + \eta \left(\frac{1}{N}\sum_{i=1}^N \frac{\partial L_i}{\partial \mathbf{\theta}} \right)$\\
$\mathbf{\theta} \leftarrow \mathbf{\theta} - \mathbf{v}_\theta$\\
            }
        }
    }
\caption{Embedding space learning through Gradient Descent}
\label{alg:learning}
\end{algorithm}

To this end, a Triplet Deep Network is designed. It consists of three base networks which share the same parameters, each taking the descriptor of a shot as input, and computing the desired embedding function $\phi(\cdot)$.
The loss of the network for a training triplet $(\mathbf{x}_i, \mathbf{x}_i^+, \mathbf{x}_i^-)$ is defined by the Hinge loss as
\ifCLASSOPTIONdraftcls
	\begin{equation}
	L_i(\mathbf{w}, \mathbf{\theta}) = \max \left( 0, \| \phi(\mathbf{x}_i) - \phi(\mathbf{x}_i^+) \|^2 + 
		\left(1 - \| \left. \phi(\mathbf{x}_i) - \phi(\mathbf{x}_i^- \right) \|^2 \right) \right)
	\end{equation}
\else
	\begin{multline}
	L_i(\mathbf{w}, \mathbf{\theta}) = \max \left( 0, \| \phi(\mathbf{x}_i) - \phi(\mathbf{x}_i^+) \|^2 + \right. \\
	+ \left. \left(1 - \| \left. \phi(\mathbf{x}_i) - \phi(\mathbf{x}_i^- \right) \|^2 \right) \right)
	\end{multline}
\fi
where $\mathbf{w}$ are the network weights, and $\mathbf{\theta}$ are biases. The overall loss for a batch of $N$ triplets is given by the average of the losses for each triplet, plus a $L_2$ regularization term on network weights to reduce over-fitting
\begin{equation}
L(\mathbf{w}, \mathbf{\theta}) = \frac{\lambda}{2} \| \mathbf{w} \|^2 + \frac{1}{N}\sum_{i=1}^N L_i(\mathbf{w}, \mathbf{\theta}).
\label{eq:loss}
\end{equation}

During learning, we perform mini-batch Stochastic Gradient Descent (SGD). At each iteration, we randomly sample $N$ training triplets. For every triplet, we calculate the gradients over its components and perform back propagation according to Eq.~\ref{eq:loss}. Details of the learning procedure are given in Algorithm~\ref{alg:learning}.

The embedding network computes the projection $\phi(\mathbf{x})$ of a shot in the embedding space by means of three fully connected layers having, respectively, 500, 125 and 30 neurons, with ReLU activation. These are interleaved with Dropout layers~\cite{srivastava2014dropout}, with retain probability 0.5, to reduce over-fitting. Since the embedding network is replicated three times to compute the final Triplet loss, Dropout is synchronized among the three branches, so that the same neurons are deactivated when computing $\phi(\mathbf{x}_i)$, $\phi(\mathbf{x}_i^+)$ and $\phi(\mathbf{x}_i^-)$.

The overall network is trained with momentum $\gamma=0.9$ and regularization strength $\lambda = 0.0005$. The learning rate $\eta$ is initially set to 0.01 and then scaled to 0.001 after 50 iterations. Training is performed in mini-batches containing $N=500$ triplets. The amount of regularization and number of neurons were selected with a grid search on the BBC Planet Earth dataset, the most challenging we used.

\subsection{Temporal Aware Clustering}
\label{sec:temporal_aware_clustering}

To obtain a temporal segmentation of the video we require segments to be as semantically homogeneous as possible. Inspired by k-means, a cluster homogeneity may be described by the sum of squared distances between cluster elements and its centroid, called within-group sum of squares (WSS). A reasonable objective is thus minimizing the total within-group sum of squares (TWSS), i.e. the sum of the WSS for all clusters. Differently from k-means, we would also like to find the number of clusters, with the additional constraint of them being temporally continuous intervals. Minimizing the TWSS alone, would lead to the trivial solution of having a single shot in each sequence, so a penalty term needs to be added to avoid over-segmentation.

The problem we need to solve is thus
\begin{equation}
\label{eq:kts}
\min_{m, t_1, ..., t_{m}} \sum_{i=0}^{m} \mli{WSS}_{t_{i},t_{i+1}} + Cg(m,n) 
\end{equation}
where $m$ is the number of change points by which the input video is segmented, $t_i$ is the position of $i$-th change point ($t_0$ and $t_{m+1}$ are the beginning and the end of the video respectively), and $\mli{WSS}_{t_i,t_{i+1}}$ is the within-group sum of squares of the $i$-th segment in the embedding space. The term $g(m,n) = m(\log(n/m)+1)$ is a Bayesian information criterion penalty~\cite{friedman2001elements} parametrized with the number of segments $m$ and the number of shots in the video $n$, which aims to reduce the over-segmentation effect. Parameter $C$ tunes the relative importance of the penalty: higher values of $C$ penalize segmentations with too many segments.

The sum of squared distances between a set of points and their mean can be expressed as a function of the pairwise squared distances between the points alone. Therefore, the within-group sum of squares can be written as
\ifCLASSOPTIONdraftcls
\begin{equation}
\mli{WSS}_{t_i,t_{i+1}} \triangleq \sum_{t=t_i}^{t_{i+1}-1} \| \phi(\mathbf{x}_t) - \mathbf{\mu}_{i} \|^2 = \frac{1}{2(t_{i+1}-t_i)} \sum_{i,j=t_i}^{t_{i+1}-1} \| \phi(\mathbf{x}_i) - \phi(\mathbf{x}_j) \|^2
\end{equation}
\else
\begin{align}
\mli{WSS}_{t_i,t_{i+1}} & \triangleq \sum_{t=t_i}^{t_{i+1}-1} \| \phi(\mathbf{x}_t) - \mathbf{\mu}_{i} \|^2 \nonumber \\
& = \frac{1}{2(t_{i+1}-t_i)} \sum_{i,j=t_i}^{t_{i+1}-1} \| \phi(\mathbf{x}_i) - \phi(\mathbf{x}_j) \|^2
\end{align}
\fi
where $\mu_i$ is the mean of each story, defined as:
\begin{equation}
\mathbf{\mu}_i = \frac{1}{t_{i+1}-t_i}\sum_{t=t_i}^{t_{i+1}-1} \phi(\mathbf{x}_t)
\end{equation}

The temporal clustering objective (Eq.~\ref{eq:kts}) can, in this way, be minimized using a Dynamic Programming approach. First, $\mli{WSS}_{k,k+d}$ is computed for each possible starting point $k$ and segment duration $d$. Then, the objective is minimized by iteratively computing the best objective value for the first $j\in[1,n]$ shots and $m\in[0,n-1]$ change points
\begin{equation}
D_{m,j} = \min_{k=m,...,j-1}(D_{m-1,k}+\mli{WSS}_{k,j})
\end{equation}
having set $D_{0,j}=\mli{WSS}_{0,j}$.

The optimal number of change points is then selected as $m^* = \argmin_{m} D_{m,n} + Cg(m,n)$, and the best segmentation into stories is reconstructed by backtracking.

\section{Story presentation with aesthetically pleasing thumbnails}
\label{sec:retrieval}
\begin{figure*}[tb]
	\def \thumbwidth {0.11\textwidth}
    \centering
    \subfigure[Input image] {
        \includegraphics[width=\thumbwidth]{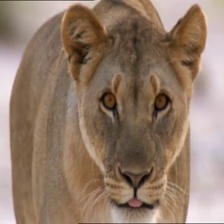}
    }
    \subfigure[\texttt{conv1*}] {
        \includegraphics[width=\thumbwidth]{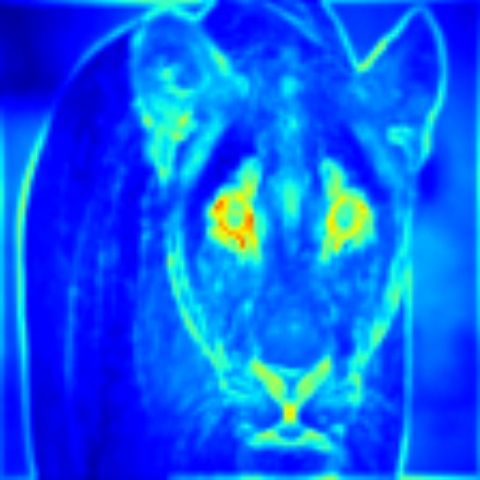}
    }
    \subfigure[\texttt{conv2*}] {
        \includegraphics[width=\thumbwidth]{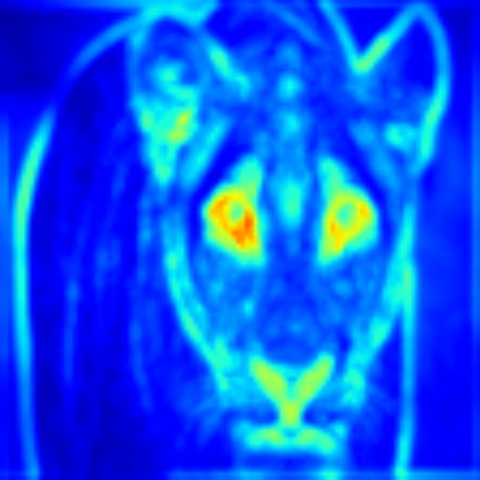}
    }
    \subfigure[\texttt{conv3*}] {
        \includegraphics[width=\thumbwidth]{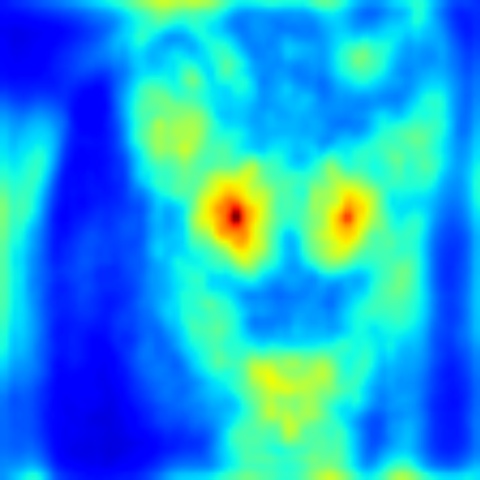}
    }
    \subfigure[\texttt{conv4*}] {
        \includegraphics[width=\thumbwidth]{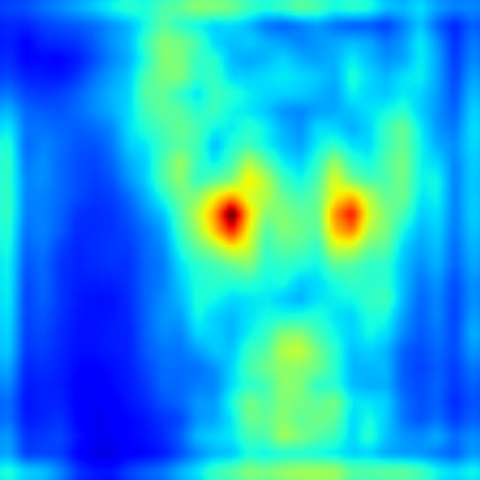}
    }
    \subfigure[\texttt{conv5*}] {
        \includegraphics[width=\thumbwidth]{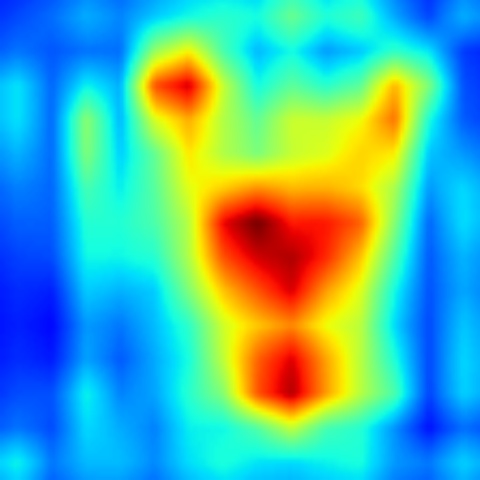}
    }\\
    
    \subfigure[Input image] {
        \includegraphics[width=\thumbwidth]{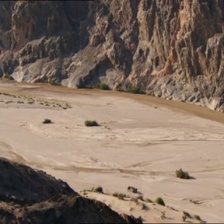}
    }
    \subfigure[\texttt{conv1*}] {
        \includegraphics[width=\thumbwidth]{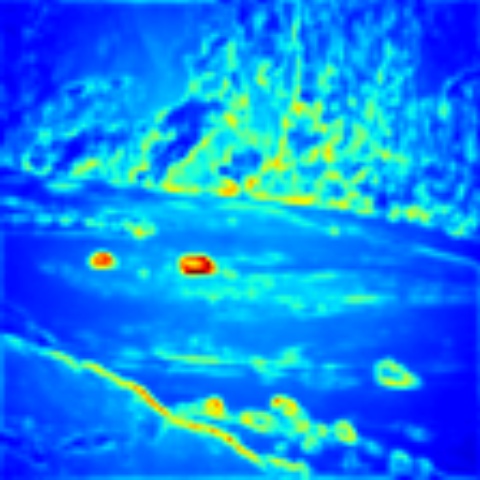}
    }
    \subfigure[\texttt{conv2*}] {
        \includegraphics[width=\thumbwidth]{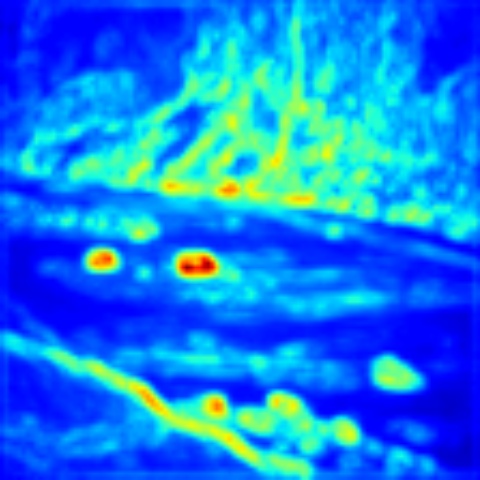}
    }
    \subfigure[\texttt{conv3*}] {
        \includegraphics[width=\thumbwidth]{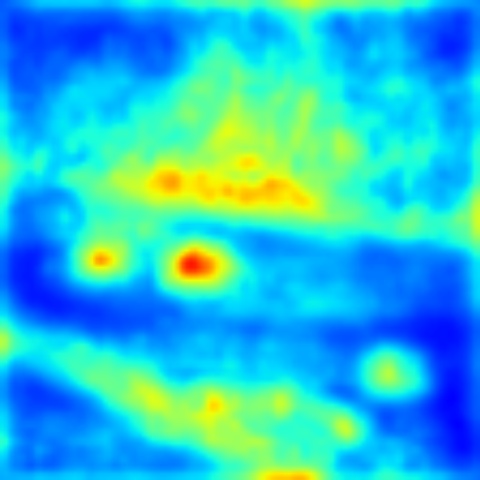}
    }
    \subfigure[\texttt{conv4*}] {
        \includegraphics[width=\thumbwidth]{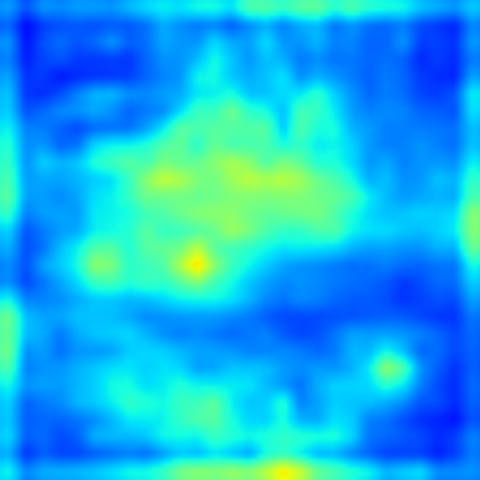}
    }
    \subfigure[\texttt{conv5*}] {
        \includegraphics[width=\thumbwidth]{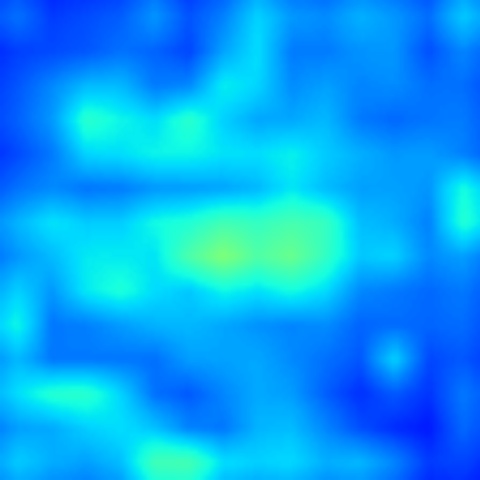}
    } 
    \caption{Hypercolumn features extracted from two sample images. Each map represents the mean activation map over a set of layers: (b) and (h) are built using layers \texttt{conv1\_1} and \texttt{conv1\_2}, (c) and (i) with layers \texttt{conv2\_1} and \texttt{conv2\_2}; (d) and (j) with \texttt{conv3\_1}, \texttt{conv3\_2} and \texttt{conv3\_3}; (e) and (k) with \texttt{conv4\_1}, \texttt{conv4\_2}, and \texttt{conv4\_3}. Finally, (f) and (i) are built using layers \texttt{conv5\_1}, \texttt{conv5\_2} and \texttt{conv5\_3}. Best viewed in color.}
    \label{fig:hypercolumns}
\end{figure*}

The availability of the video structure, i.e. its layered decomposition in stories, shots and keyframes, is not just an indexing tool for easier navigation or section selection, but may be employed as an extremely effective presentation aid. Given a set of videos relevant to a query term $q$, we can leverage the story structure to point to the most relevant part of the video and use the two lower layers (shots and keyframes) to cheaply select an aesthetically pleasing and semantically significant presentation.

For each relevant video, we build a ranking function which returns an ordered set of (video, story, thumbnail) triplets. In each triplet, the retrieved story must belong to the retrieved video, and should be as consistent as possible with the given query. Moreover, the returned thumbnail must belong to the given story, and should be representative of the query as well as aesthetically remarkable.


Given a query $q$, we first match $q$ with the most similar detected concept $u$, using the Word2Vec embedding. If the query $q$ is composed by more than one word, the mean of the embedded vectors is used. The probability function $P(s,u)$, defined in Eq.~\ref{eq:proba_unigram}, accounts for the presence of a particular unigram in one shot, and is therefore useful to rank stories given a user query. Each story $a$ inside the relevant set is then assigned a score according to the following function:
\begin{equation}
R_{a}(q) = \max_{s} \left(\alpha P(s,u) + (1-\alpha) \max_{d \in s} A(d)\right)
\label{eq:retrieval}
\end{equation}
where $s$ is a shot inside the given story, and $d$ represents a keyframe extracted from a given shot. Parameter $\alpha$ tunes the relative importance of semantic representativeness with respect to function $A(d)$, which is a measure of the aesthetic beauty. The final retrieval results is a collection of stories, ranked according to $R_{a}(q)$, each one represented with the keyframe that maximizes the second term of the score.

\subsection{Thumbnail selection}
In order to evaluate how much aesthetically pleasing a thumbnail is, we should account for low level characteristics, like color, edges and sharpness, as well as high level features, such as the presence of a clearly visible and easily recognizable object. We claim that the need of low and high level features is an excellent match with the hierarchical nature of CNNs: convolutional filters, indeed, are known to capture low level as well as high level characteristics of the input image. This has also been proved by visualization and inversion techniques, like~\cite{zeiler2014visualizing} and~\cite{mahendran2015understanding}.

Being activations from convolutional filters discriminative for visual representativeness, a ranking strategy could be set up to learn their relative importance given a dataset of user preferences. However, medium sized CNNs, like the VGG-16 model, contain more than 4000 convolutional filters: this makes the use of raw activations infeasible with small datasets. Moreover, maps from different layers have different sizes, due to the presence of pooling layers. To overcome this issue, we resize each activation map to fixed size with bilinear interpolation, and average feature maps coming from the different layers, inspired by the Hypercolumn approach presented in~\cite{hariharan2015hypercolumns}. Since the user usually focuses on the center of the thumbnail rather than its exterior, each maps is multiplied by a normalized gaussian density map, centered on the center of the image and with horizontal and vertical standard deviations equal to $\sigma_b\cdot l$, where $l \times l$ is the size of the CNN input. Parameter $\sigma_b$ was set to 0.3 in all our experiments.

Following the VGG-16 architecture~\cite{simonyan2014very}, we build five hypercolumn maps, each one summarizing convolutional layers before each pooling layer: the first one is computed with activation maps from layers \texttt{conv1\_1} and \texttt{conv1\_2}; the second one with \texttt{conv2\_1} and \texttt{conv2\_2}; the third with \texttt{conv3\_1}, \texttt{conv3\_2} and \texttt{conv3\_3}; the fourth with \texttt{conv4\_1}, \texttt{conv4\_2} and \texttt{conv4\_3}; the last with \texttt{conv5\_1}, \texttt{conv5\_2} and \texttt{conv5\_3}. An example of the resulting activation maps is presented in Fig.~\ref{fig:hypercolumns}: as it can be seen, both low level and high level layers are useful to distinguish between a significant and non significant thumbnail.

To learn the relative contribution of each hypercolumn map, we rank thumbnails from each story according to their visual representativeness, and learn a linear ranking model. Given a dataset of stories $\{ a_i \}_{i=0}^m$, each with a ranking $r_i^*$, expressed as a set of pairs $(d_i, d_j)$, where thumbnail $d_i$ is annotated as more relevant than thumbnail $d_j$, we solve the following problem:
\begin{equation}
\underset{\mathbf{w_r}, \mathbf{\epsilon}}{\min} \ \frac{1}{2} \|\mathbf{w_r}\|^2 + C_r\sum_{i,j,k}\epsilon_{i,j,k} \\
\label{eq:svm-rank}
\end{equation}
subject to
\begin{equation}
\begin{array}{ll}
\forall (d_i,d_j) \in r_1^* : \mathbf{w_r}\tau(d_i) \geq \mathbf{w_r}\tau(d_j) + 1 - \epsilon_{i,j,1} \\
\ldots \\
\forall (d_i,d_j) \in r_m^* : \mathbf{w_r}\tau(d_i) \geq \mathbf{w_r}\tau(d_j) + 1 - \epsilon_{i,j,m} \\
\forall i, j, k : \epsilon_{i,j,k} \geq 0
\end{array}
\end{equation}
where $\tau(d_i)$ is the feature vector of thumbnail $d_i$, which is composed by the mean and standard deviation of each hypercolumn map extracted from the thumbnail itself. $C_r$ allows trading-off the margin size with respect to the training error. The objective stated in Eq.~\ref{eq:svm-rank} is convex and equivalent to that of a linear SVM on pairwise difference vectors $\tau(d_i)-\tau(d_j)$~\cite{joachims2002optimizing}. The final aestethic score for keyframe $d$ is given by $A(d)=\mathbf{w_r}\tau(d)$.

\section{Dealing with subjectivity}
\label{sec:subjectivity}
\subsection{Evaluation protocol}
Measuring story detection performance is significantly different from measuring shot detection performance. Indeed, classical boundary detection scores, such as Precision and Recall, fail to convey the true perception of an error, which is different for an off-by-one shot or for a completely missed story boundary.

Better fitting measures were proposed in~\cite{vendrig02}: Coverage measures the quantity of shots belonging to the same story correctly grouped together, while Overflow evaluates to what extent shots not belonging to the same story are erroneously grouped together. 
%
%
An F-Score measure, $F_{co}$, can be defined to combine Coverage and Overflow in a single measure, by taking the harmonic mean of Coverage and 1-Overflow.
These measures are nevertheless known to have some drawbacks, which may affect the evaluation. As also noted in~\cite{sidiropoulos12}, $F_{co}$ is not symmetric, leading to unusual phenomena in which an early or late positioning of the story boundary, of the same amount of shots, may lead to strongly different results. Moreover, the relation of Overflow with the previous and next stories creates unreasonable dependencies between an error and the length of a story observed many shots before it.

An alternative symmetric measure, based on intersection over union, was proposed in \cite{acm}, and was proved to be more effective. Here, a story in a video is represented as a closed interval, where the left bound of the interval is the starting frame of the story, and the right bound is the ending frame of the sequence. The intersection over union of two stories $a$ and $b$, $\text{IoU}(a,b)$, can therefore be written as
\begin{equation}
\text{IoU}(a,b) = \frac{a \cap b}{a \cup b}
\end{equation}

A segmentation of a video into stories can be seen as a set of non-overlapping stories, whose union is the set of frames of the video. By exploiting this relation, \cite{acm} defines the intersection over union of two segmentations $\mathcal{A}$ and $\mathcal{B}$ as
\ifCLASSOPTIONdraftcls
	\begin{equation}
		\overline{\text{IoU}}(\mathcal{A},\mathcal{B}) = \frac{1}{2} \left(
		\frac{1}{\#\mathcal{A}} \sum_{a \in \mathcal{A}} \max_{b \in \mathcal{B}} \text{IoU}(a,b) + 
		\frac{1}{\#\mathcal{B}} \sum_{b \in \mathcal{B}} \max_{a \in \mathcal{A}} \text{IoU}(a,b)  \right)
		\label{eq:iou}
	\end{equation}
\else
	\begin{multline}
		\overline{\text{IoU}}(\mathcal{A},\mathcal{B}) = \frac{1}{2} \left(
		\frac{1}{\#\mathcal{A}} \sum_{a \in \mathcal{A}} \max_{b \in \mathcal{B}} \text{IoU}(a,b) + \right. \\
		\left. +\frac{1}{\#\mathcal{B}} \sum_{b \in \mathcal{B}} \max_{a \in \mathcal{A}} \text{IoU}(a,b)  \right)
		\label{eq:iou}
	\end{multline}
\fi

It is easy to see that, considering the particular case of $\mathcal{A}$ being the ground-truth annotation and $\mathcal{B}$ being the segmentation produced by an algorithm, Eq.~\ref{eq:iou} computes, for each ground-truth story, the maximum intersection over union with the detected stories. Then, the same is done for detected stories against ground-truth ones, and the two quantities are averaged.

\subsection{Finding an agreement between annotations}
\label{sec:agreement_dp}
Being story detection a considerably subjective task, it is often the case that the same video is annotated differently by more than one annotator. This results in a set of annotations for each video, while an automatic model should produce a single segmentation, as consistent as possible with all given human annotations. Therefore, the prediction of a model should not be compared with each given annotation, but with the segmentation which is most similar to all given annotations.

In this paragraph, we investigate the problem of finding the segmentation which maximizes the agreement with respect to a set of annotations. Formally, given a set of $m$ annotations $\mathbb{S}$, we aim at finding the segmentation $\mathcal{A}^*$ which maximizes the average intersection over union with respect to $\mathbb{S}$
\begin{equation}
    \mathcal{A}^* = \argmax_\mathcal{A} \frac{1}{m}\sum_{\mathcal{S} \in \mathbb{S}} \overline{\text{IoU}}(\mathcal{A}, \mathcal{S})
    \label{eq:obj_dp}
\end{equation}

Algebraic manipulation reveals that given Eq.~\ref{eq:iou}, this maximization is equivalent to find $\mathcal{A}$ to maximize
\ifCLASSOPTIONdraftcls
	\begin{equation}
	J(\mathcal{A}) = \underbrace{ 
	    \frac{1}{\# \mathcal{A}} \sum_{a_i \in \mathcal{A}} \sum_{\mathcal{S} \in \mathbb{S}} \max_{s_j \in \mathcal{S}} \left(\text{IoU}(a_i,s_j)\right) 
	    }_{J_1(\mathcal{A})} + \underbrace{
	    \sum_{\mathcal{S} \in \mathbb{S}} \frac{1}{\# \mathcal{S}} \sum_{s_j \in \mathcal{S}} \max_{{a_i} \in \mathcal{A}} \left(\text{IoU}(a_i, s_j) \right) 
	    }_{J_2(\mathcal{A})}
	\label{eq:J}
	\end{equation}
\else
	\begin{multline}
	J(\mathcal{A}) = \underbrace{ 
	    \frac{1}{\# \mathcal{A}} \sum_{a_i \in \mathcal{A}} \sum_{\mathcal{S} \in \mathbb{S}} \max_{s_j \in \mathcal{S}} \left(\text{IoU}(a_i,s_j)\right) 
	    }_{J_1(\mathcal{A})} + \\
	    + \underbrace{
	    \sum_{\mathcal{S} \in \mathbb{S}} \frac{1}{\# \mathcal{S}} \sum_{s_j \in \mathcal{S}} \max_{{a_i} \in \mathcal{A}} \left(\text{IoU}(a_i, s_j) \right) 
	    }_{J_2(\mathcal{A})}
	\label{eq:J}
	\end{multline}
\fi

Given a video with $n$ shots, a brute force resolution of Eq.~\ref{eq:obj_dp} would require to test every possible annotation compatible with the shots, thus leading to a time complexity $O(2^n)$.

We approximate the maximization of Eq. \ref{eq:J} as a longest-path problem in a graph. Let $\mathcal{G}(V, E)$ be a weighted directed acyclic graph with vertices representing shots of the video, $V = \{ 1, 2, ..., n\}$, and edges connecting each vertex to all the other nodes with greater values, $E = \{ (i, j): i, j \in V, i < j\}$. A valid segmentation of the video can be seen as a path $\mathcal{P}$ in $\mathcal{G}$ having source node 1 and target node $n$. In this configuration, each edge $(i,j)$ in the path corresponds to a story in the segmentation.

If the number of stories in the segmentation (i.e. its cardinality) is known in advance, then $J_1(\mathcal{A})$ can be decomposed into a sum of edge weights, as follows:
\begin{equation}
    J_1(\mathcal{A}) = \sum_{(i,j) \in \mathcal{P}} w_1(i,j)
\end{equation}
where the weight of an edge $(i,j)$ is
\begin{equation}
    w_1(i,j) = \frac{1}{l} \sum_{\mathcal{S} \in \mathbb{S}} \max_{s_k \in \mathcal{S}} \left(\text{IoU}(\left[ i, j\right],s_k)\right) 
\end{equation}
$\left[ i, j\right]$ is the story corresponding to edge $(i,j)$, and $l$ is the length of the segmentation (indicated as $\# \mathcal{A}$ in Eq.~\ref{eq:J}).

Unfortunately, $J_2(\mathcal{A})$ cannot be factored in the same way. However, we notice that it can be rewritten as follows:
\begin{equation}
    J_2(\mathcal{A}) = \sum_{t=1}^{\# \mathcal{P}} w_2(\mathcal{P}, t)
\end{equation}
where
\ifCLASSOPTIONdraftcls
	\begin{equation}
		w_2(\mathcal{P}, t) = \left(\sum_{\mathcal{S} \in \mathbb{S}} \frac{1}{\#\mathcal{S}} 
			\sum_{s_j \in \mathcal{S}} \max_{{a_i} \in \mathcal{A}_t} \left(\text{IoU}(a_i, s_j) \right)  
			- \sum_{\mathcal{S} \in \mathbb{S}} \frac{1}{\#\mathcal{S}} 
			\sum_{s_j \in \mathcal{S}} \max_{{a_i} \in \mathcal{A}_{t-1}} \left(\text{IoU}(a_i, s_j) \right) \right)
	\end{equation}
\else
	\begin{multline}
		w_2(\mathcal{P}, t) = \left(\sum_{\mathcal{S} \in \mathbb{S}} \frac{1}{\#\mathcal{S}} 
			\sum_{s_j \in \mathcal{S}} \max_{{a_i} \in \mathcal{A}_t} \left(\text{IoU}(a_i, s_j) \right)  + \right. \\
		\left. - \sum_{\mathcal{S} \in \mathbb{S}} \frac{1}{\#\mathcal{S}} 
			\sum_{s_j \in \mathcal{S}} \max_{{a_i} \in \mathcal{A}_{t-1}} \left(\text{IoU}(a_i, s_j) \right) \right)
	\end{multline}
\fi
where $\mathcal{A}_t$, at each step $t$ of the path, is the set of stories corresponding to already visited nodes.

The maximization of $ J_1 + J_2$ can be addressed as the problem of finding the longest path of length $l$ in $\mathcal{G}$, and approximately solved through a Dynamic Programming strategy, by pretending that $J_2(\mathcal{A})$ is a sum of edge weights (even though $w_2(\mathcal{P}, t)$ actually depend on the specific path).

Having chosen a path length $l$, For each $1 \leq i \leq l$, and every vertex $v$, we compute $D[i,v]$ where $D[i,v]$ is the weight of the longest walk of length exactly $i$ starting at vertex 1 and ending at vertex $v$. To compute $D[l,n]$, we use the following relation:
\begin{equation}
    D[i+1, v] = \max_{x \in Pred(v)} \left( D[i, x] + w_1(\mathcal{P}, t) + w_2(x,v)  \right)
\end{equation}
where $Pred(v)$ is the predecessor set of vertex $v$, and $w_1(\mathcal{P}, t)$ is computed by considering the path used in $D[i, x]$, plus node $v$. The best path from vertex 1 to vertex $n$ with length $l$ is then reconstructed by backpropagation, and the same procedure is repeated for $1 \leq l \leq n$. $\mathcal{A}^*$ is then selected as the path of maximum cost.

Since for each $l$ the Dynamic Programming algorithm has time complexity $O(l\cdot n)$, the overall complexity is $O(n^3)$, being $n$ the number of shots in the video.

It is worth mentioning that to assess the quality of the proposed approximation, we tested it on 11.000 randomly generated sequences for which $\mathcal{A}^*$ has been computed with brute-force, with length $n = 100$, a number of stories varying from 2 and 7, and with a number of annotations $m$ ranging from 2 to 10. 98.4\% of the generated segmentations where correct, while the mean absolute error, in terms of $\overline{\text{IoU}}$, was $1.16\cdot 10^{-5}$.

\section{Experimental evaluation}
\label{sec:experimental}
We compare our story detection approach against state of the art algorithms from the literature which are applicable, and perform experiments to assess the role of the proposed features and embedding. In addition, we address the subjective nature of story detection by tackling our embedding to learn the style of different annotators, and the segmentation provided by the algorithm described in Section~\ref{sec:agreement_dp}. Finally, we evaluate the effectiveness of the proposed video retrieval strategy, both quantitatively and qualitatively.

To perform shot detection, we use an off-the-shelf shot detector~\cite{apostolidis14} which relies on SURF descriptors and HSV color histograms. Abrupt transitions are detected by thresholding a distance measure between frames, while longer gradual transitions are detected by means of the derivative of the moving average of the aforesaid distance.

\begin{figure}
\centering
\includegraphics[width=0.95\columnwidth]{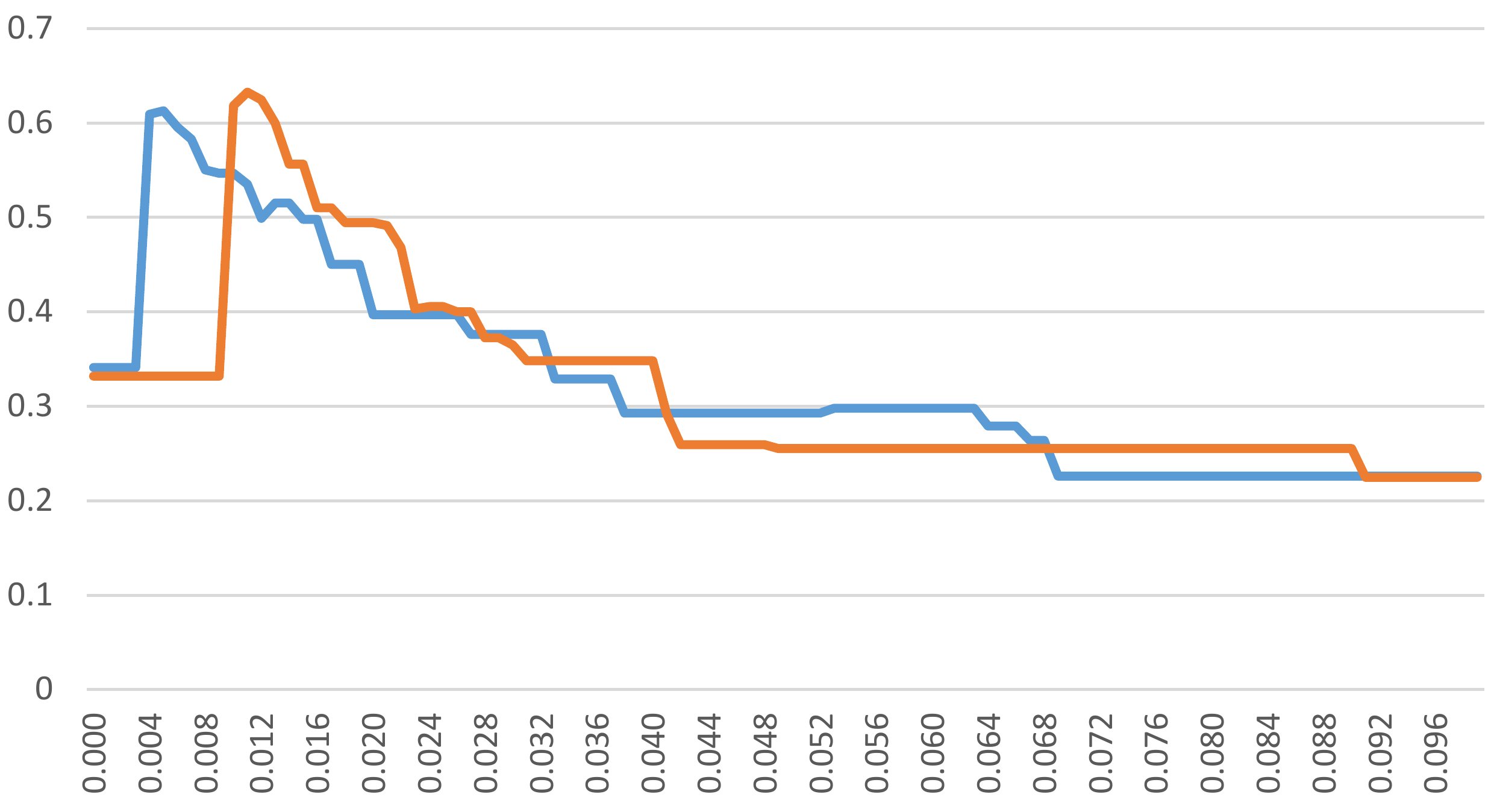}
\caption{Parameter $C$ influence. Variation of $\overline{\text{IoU}}$ with respect to $C$ for two different videos of BBC Planet Earth.}
\label{fig:C}
\end{figure}

\subsection{Datasets}
To test the temporal segmentation capabilities of our model, we run a series of experimental tests on the Ally McBeal dataset released in~\cite{ercolessi2011segmenting}, which contains the temporal segmentation into stories of four episodes of the first season of Ally McBeal. The dataset contains 2660 shots and 160 stories. Closed captions were used as transcript.

We also employ the BBC Planet Earth dataset~\cite{acm}, which contains the segmentation into stories of eleven episodes from the BBC documentary series \textit{Planet Earth}~\cite{bbcpe}. Each episode is approximately 50 minutes long, and the whole dataset contains around 4900 shots and 670 stories. Each video is also provided with the corresponding transcript. To augment the dataset, and test the proposed way to deal with different annotations, we asked four more annotators to segment each video in the dataset.

It is worth to mention that the aforementioned datasets are considerably different, both because of the nature of the videos they contain, and because of the kind of annotation. Indeed, the annotation in Ally McBeal reproduces the partitioning of a TV series into stories, which is mainly based on the dialogues and the location of the stories, while the annotation of the BBC Planet Earth episodes is far more difficult to reproduce, since it relies on the semantics of the video and of the speaker transcript.

\begin{table}[tb]
\begin{center}
\caption{Story detection performance on the Ally McBeal dataset, and previous methods vs. our approach.}
\begin{tabular}{|c|c|c|c|c|} 
\hline
\textbf{Episode} & STG \cite{sidiropoulos11} & NW \cite{chasanis09} & SDN \cite{acm} & Ours \\  \hline
Ep. 1 		& 0.65 & 0.38 & 0.37 & \textbf{0.98} \\ \hline
Ep. 2 		& 0.70 & 0.36 & 0.34 & \textbf{0.86} \\ \hline
Ep. 3 		& 0.72 & 0.40 & 0.36 & \textbf{0.94} \\ \hline
Ep. 4 		& 0.63 & 0.36 & 0.30 & \textbf{0.96} \\ \hline
\hline         
\textbf{Average}        & 0.68 & 0.38 & 0.34 & \textbf{0.94}         \\  \hline 
\end{tabular}
\label{tab:results_ally}
\end{center}
\end{table}

\begin{table}[tb]
\begin{center}
\caption{Story detection performance on the individual episodes from BBC Planet Earth, and previous methods vs. our approach.}
\begin{tabular}{|c|c|c|c|c|} 
\hline
\textbf{Episode} & STG \cite{sidiropoulos11} & NW \cite{chasanis09} & SDN \cite{acm} & Ours 	\\  \hline
From Pole to Pole       & 0.42          & 0.35          & 0.50          & \textbf{0.72}          \\  \hline 
Mountains               & 0.40          & 0.31          & 0.53          & \textbf{0.75}          \\  \hline 
Fresh Water             & 0.39          & 0.34          & 0.52          & \textbf{0.67}          \\  \hline 
Caves                   & 0.37          & 0.33          & 0.55          & \textbf{0.62}          \\  \hline 
Deserts                 & 0.36          & 0.33          & 0.36          & \textbf{0.62}          \\  \hline 
Ice Worlds              & 0.39          & 0.37          & 0.51          & \textbf{0.73}          \\  \hline 
Great Plains            & 0.46          & 0.37          & 0.47          & \textbf{0.63}          \\  \hline 
Jungles                 & 0.45          & 0.38          & 0.51          & \textbf{0.62}          \\  \hline 
Shallow Seas            & 0.46          & 0.32          & 0.51          & \textbf{0.74}          \\  \hline 
Seasonal Forests        & 0.42          & 0.20          & 0.38          & \textbf{0.65}          \\  \hline 
Ocean Deep              & 0.34          & 0.36          & 0.48          & \textbf{0.65}          \\  \hline 
\hline         
\textbf{Average}        & 0.41          & 0.33          & 0.48          & \textbf{0.67}			 \\  \hline 
\end{tabular}
\label{tab:results_bbcph}
\end{center}
\end{table}

\subsection{Comparison with the State of the art}
The performance of our method depends on the selection of hyperparameter $C$ in the temporal clustering objective (Eq.~\ref{eq:kts}), which yields a trade-off between over- and under-segmentation. Figure~\ref{fig:C} reports an example of the variation of intersection over union with respect to $C$ for the two different videos of the BBC Planet Earth dataset. 
Clearly, each chart presents a global maximum, but the optimal $C$ value changes from video to video. This would lead to a sub-optimal choice of $C$, if selected with cross-validation. The temporal clustering selects a story for each shot for low values of $C$ and as soon as the parameter goes over a certain value, the clustering begins to provide very significant groupings. For this reason, our choice of $C$ is video dependent and, using a step of $0.001$, we increase the $C$ value until the number of clusters is lower than the number of shots in the video. This may be sub-optimal, but the results are totally independent of the training phase and do not require assumptions on the specific video.

Our model is compared against three recent proposals for video decomposition:~\cite{sidiropoulos11}, which uses a variety of visual and audio features merged in a Shot Transition Graph (STG);~\cite{chasanis09}, that combines low level color features with the Needleman-Wunsh (NW) algorithm, and~\cite{acm}, which exploits visual features extracted with a CNN and Bag-of-Words histograms extracted from the transcript, which are merged in a Siamese Deep Network (SDN).

We use the executable of~\cite{sidiropoulos11} and the source code of~\cite{acm} provided by the authors, and re-implement the method in~\cite{chasanis09}. Parameters of all methods were selected to maximize the performance on the training set. The shot detector we use is the same of~\cite{sidiropoulos11}, so performance results are not affected by differences in the shot detection phase.

\begin{table}[tb]
    \centering
    \caption{Evaluation on the Ally McBeal dataset, when training on BBC Planet Earth and on Ally McBeal.}

    \begin{tabular}{|l|c|c|}
        \hline
        \textbf{Episode} & \textbf{Train on BBC PE} & \textbf{Train on AMB} \\
        \hline
        Ep. 1 & 0.87 & 0.98 \\ \hline
        Ep. 2 & 0.81 & 0.86 \\ \hline
        Ep. 3 & 0.93 & 0.94 \\ \hline
        Ep. 4 & 0.91 & 0.96 \\ \hline
        \hline
        \textbf{Average} & \textbf{0.88} & \textbf{0.94} \\ \hline
    \end{tabular} 
    \label{tab:ally}
\end{table}

In Tables~\ref{tab:results_ally} and~\ref{tab:results_bbcph} we compare the performance of our method with respect to the aforementioned methods, on Ally McBeal and BBC Planet Earth, using annotations provided in~\cite{acm}. All experiments were conducted in a leave-one-out setup, using one video for testing and all other videos from the same dataset as training. Reported results suggest that our embedding strategy is able to deal effectively with different kind of videos and of annotations, learning the specific annotation style of each dataset. On all datasets, indeed, our method outperforms all the approaches it has been compared to. 

To test the generality of the learned embedding, we also perform a second experiment, in which we train a model on the entire BBC Planet Earth dataset, and test it on the Ally McBeal series. The objective of the experiment is therefore to investigate how a model learned on a particular kind of videos can generalize to another category. Results are shown in Table~\ref{tab:ally}: even if the embedding has been learned on documentaries, and even if in this case visual semantic features are less effective, the model is still able to generalize to unseen kinds of videos.

\begin{table}[tb]
\begin{center}
\caption{Story detection performance with various features.}
\begin{tabular}{|c|c|c|} 
\hline
\textbf{Features/Embedding} & \textbf{Ally McBeal} & \textbf{BBC Planet Earth} \\ \hline
VA 				& 0.898 & 0.638 \\ \hline
VA+A 			& 0.915 & 0.654 \\ \hline
VA+A+QoS 		& 0.914 & 0.656 \\ \hline
VA+A+QoS+T 		& 0.921 & 0.657 \\ \hline
VA+A+QoS+T+VS 	& 0.925 & 0.660 \\ \hline
VA+A+QoS+T+VS+TS& 0.935 & 0.672 \\ \hline
\end{tabular}
\label{tab:results_features}
\end{center}
\end{table}

\begin{table}[t]
\begin{center}
\caption{Story detection performance with different embeddings.}
\begin{tabular}{|c|c|c|} 
\hline
\textbf{Embedding} & \textbf{Ally McBeal} & \textbf{BBC Planet Earth} \\ \hline
LSTM 	& 0.82 & 0.58 \\ \hline
Siamese & 0.87 & 0.49 \\ \hline
Triplet & 0.94 & 0.67\\ \hline
\end{tabular}
\label{tab:results_embeddings}
\end{center}
\end{table}

\subsection{Feature and embedding comparisons}
To test the role of the proposed features and embedding, we conducted two additional tests. In the first one, whose results are reported in Table~\ref{tab:results_features} the triplet embedding is trained using an increasing set of features: visual appearance (VA), Audio (A), Quantity of Speech (QoS), Time (T), Visual and Textual semantic (VS, TS). Results are reported in terms of mean $\overline{\text{IoU}}$. Each feature, when added, resulted in a performance improvement.

In the second experiment, we use all features and test different embeddings. We test a Siamese network with the same architecture and the same number of neurons of the Triplet network. We also train a LSTM network: the descriptor of each shot is fed to a fully connected network with the same structure of the embedding network, and then to an LSTM layer with memory size 10 and output size 1. The network is trained to predict, at each time step, the presence of a story boundary, with a binary crossentropy loss. Results, reported in Table~\ref{tab:results_embeddings} show that the proposed Triplet strategy is superior both to the Siamese and the LSTM approach. In conclusion, all features are important but the embedding architecture boosts performances.

\subsection{Feature importance analysis}
\label{sec:feat_importance}
\begin{figure}[tb]
\centering
\subfigure[Ally McBeal~\cite{ercolessi2011segmenting}] {
	\includegraphics[width=0.45\linewidth]{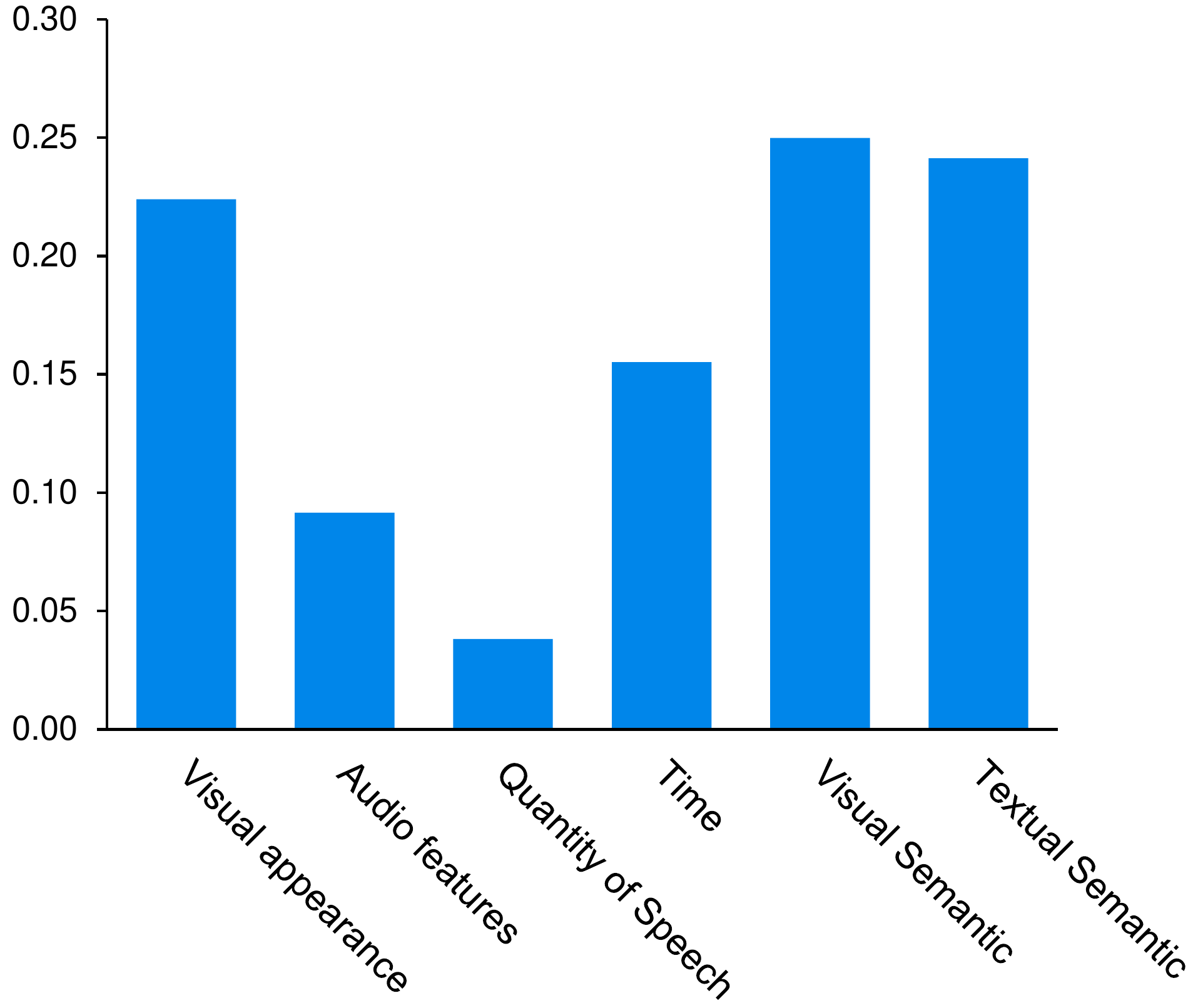}
}
\subfigure[BBC Planet Earth~\cite{acm}] {
	\includegraphics[width=0.45\linewidth]{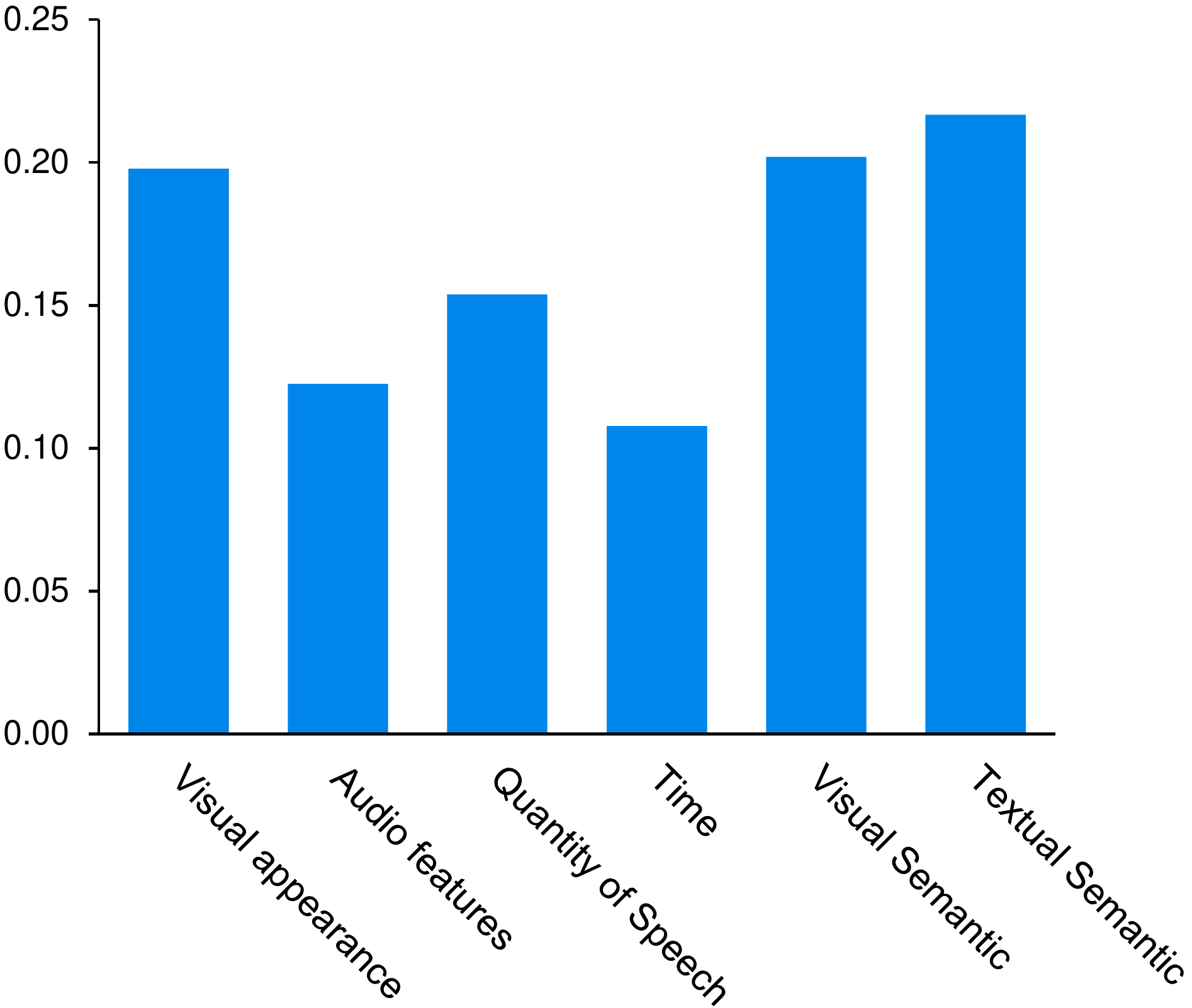}
}
\caption{Feature importance analysis. For each dataset, the relative importance of each feature is reported. See Section~\ref{sec:feat_importance} for details.}
\label{fig:importance_bph}
\end{figure}
We evaluate the relative importance and effectiveness of each of the proposed features in the final embedding. In the following, we will define the importance of a feature as the extent to which a variation of the feature can affect the embedding. Consider, for example, a linear embedding model, in which each dimension of the embedding, $\phi_i$, is given by the following equation
\begin{equation}
\phi_i(\mathbf{x}) = w_i^T \mathbf{x} + \theta_i
\end{equation}
where $w_i$ and $\theta_i$ are respectively the weight vector and the bias for the $i$-th dimension of the embedding, while $\mathbf{x}$ is the concatenation of the proposed features. In this case it is easy to see that the magnitude of elements in $w_i$ defines the importance of the corresponding features. Each feature is indeed multiplied by a subset of the $w_i$ vector, and the absolute values in $w_i$ encode the importance of each of those features. In the extreme case of a feature which is always multiplied by 0, it is straightforward to see that that feature is ignored by the $i$-th dimension of the embedding, and has therefore no importance, while a feature with high absolute values in $w$ will have a considerable effect.

\begin{figure*}[tb]
\centering
\includegraphics[width=0.91\textwidth]{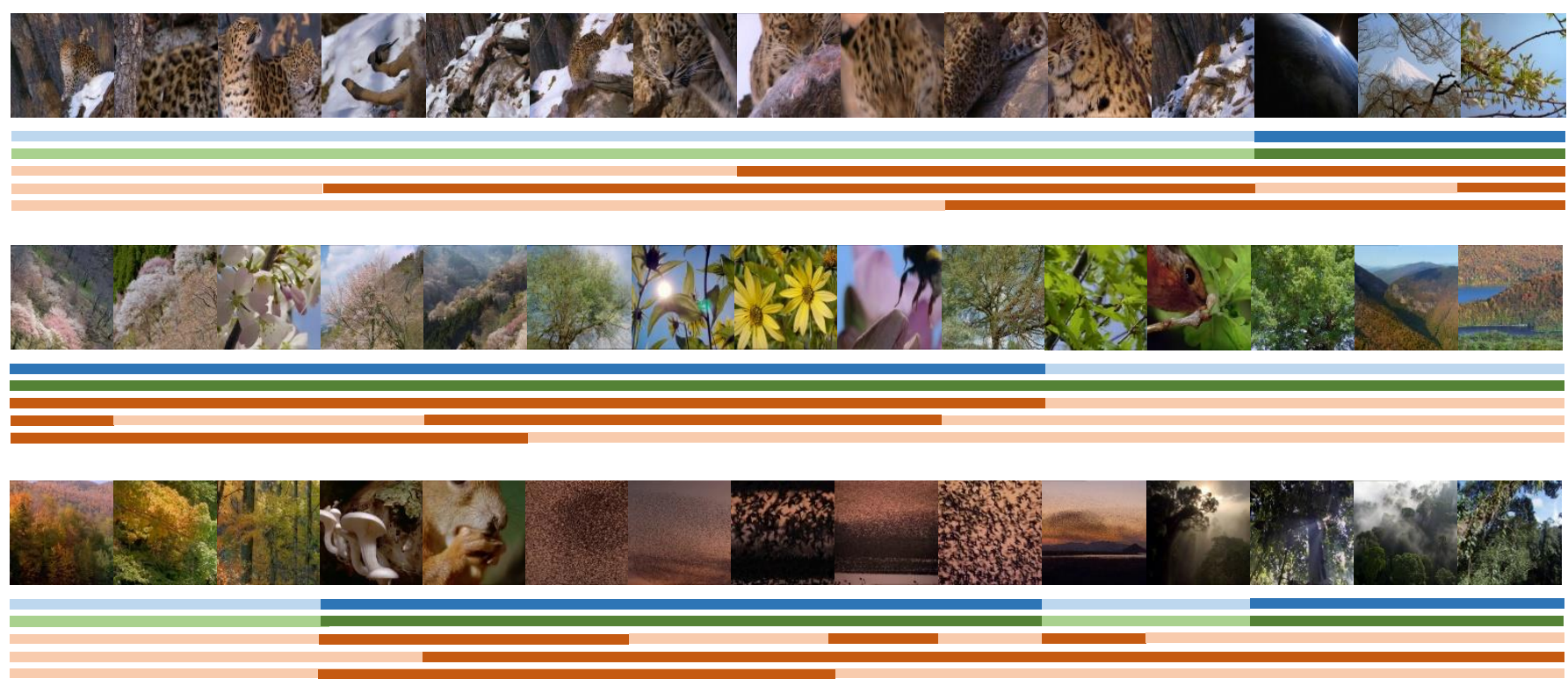}
\caption{Qualitative results on the first episode of BBC \textit{Planet Earth}. Each row represents the segmentation generated by a method, and a change in color represents a change of story. First row (blue) is the ground truth, second row (green) is our method, remaining (red) are, respectively,~\cite{sidiropoulos11},~\cite{chasanis09} and~\cite{acm} (best viewed in color).}
\label{fig:qualitative_results}
\end{figure*}

\begin{table*}[tb]
\begin{center}
\caption{Story detection performance on the BBC Planet Earth dataset, training and testing on different annotators and the maximum agreement segmentation.}
\begin{tabular}{|c|c|c|c|c|c|c|c|} 
\cline{3-8}
\multicolumn{2}{c|}{} & \multicolumn{6}{c|}{\textbf{Test on}} \\ \cline{3-8}
\multicolumn{2}{c|}{} & \textbf{Annotator 1}	& \textbf{Annotator 2}	& \textbf{Annotator 3}	& \textbf{Annotator 4}	& \textbf{Annotator 5}	& \textbf{Agreement} \\ \hline
\multirow{ 6}{*}{\rotatebox[origin=c]{90}{\textbf{Train on}}} & \textbf{Annotator 1}			& 0.669	& 0.528	& 0.428	& 0.474	& 0.416	& 0.475	\\ \cline{2-8}
& \textbf{Annotator 2}			& 0.474	& 0.654	& 0.437	& 0.505	& 0.418	& 0.541	\\ \cline{2-8}
& \textbf{Annotator 3}			& 0.455	& 0.546	& 0.572	& 0.481	& 0.404	& 0.420	\\ \cline{2-8}
& \textbf{Annotator 4}			& 0.481	& 0.536	& 0.436	& 0.606	& 0.436	& 0.433	\\ \cline{2-8}
& \textbf{Annotator 5}			& 0.468	& 0.538	& 0.432	& 0.492	& 0.545	& 0.411	\\ \cline{2-8}
& \textbf{Agreement}			& 0.605	& 0.585	& 0.547	& 0.580	& 0.454	& 0.556	\\ \hline
\end{tabular}
\label{tab:results_subjective}
\end{center}
\end{table*}

In our case, $\phi_i(\cdot)$ is a highly non-linear function of the input, thus the above reasoning is not directly applicable. Instead, given an shot $\mathbf{x}_j$, we can approximate $\phi_i(\mathbf{x}_j)$ in the neighborhood of $\mathbf{x}_j$ as follows
\begin{equation}
\phi_i(\mathbf{x}_j) \approx \nabla \phi_i(\mathbf{x}_j)^T \mathbf{x} + \theta_i
\label{eq:approx}
\end{equation}
An intuitive explanation of this approximation is that the magnitude of the partial derivatives indicates which features need to be changed to affect the embedding. Also notice that Eq.~\ref{eq:approx} is equivalent to a first order Taylor expansion.

To get an estimation of the importance of each feature regardless of the choice of $\mathbf{x}_j$, we can average the element-wise absolute values of the gradient computed in the neighborhood of each test sample

\begin{equation}
w_i = \frac{1}{N}\sum_{j=1}^N \left[ \left| \frac{\partial \phi_i}{\partial x^1}(\mathbf{x}_j) \right|, \left| \frac{\partial \phi_i}{\partial x^2}(\mathbf{x}_j) \right|, \cdots, \left| \frac{\partial \phi_i}{\partial x^d}(\mathbf{x}_j) \right| \right]
\end{equation}
where $d$ is the dimensionality of $\mathbf{x}_j$. Then, to get the relative importance of each proposed feature, we average the values of $w_i$ corresponding to that feature. The same is done for each of the dimensions of the embedding, and results are then averaged. The resulting importances for each features are finally $L_1$ normalized.

Figure~\ref{fig:importance_bph} reports the relative importance of our features on Ally Mc Beal and BBC Planet Earth. It is easy to notice that all features give a valuable contribution to the final result. 
In TV-series and documentaries visual appearance and semantic features are the most relevant cues. The quantity of speech plays an important role in documentaries, confirming that in this kind of videos a pause in the speaker discourse is often related to a story boundary, while in TV series appearance and conceptual features are often enough to perform story detection. It is also worth to notice that when the annotation to be learned is challenging, like in the BBC Planet Earth dataset, every feature becomes relevant, thus confirming the effectiveness of the proposed features.

\subsection{Qualitative results}
To give a qualitative indication of the results, in Figure~\ref{fig:qualitative_results} we report the temporal segmentation provided by our method an all the methods we compare to, as well as the ground truth annotation, on a part of the first episode of BBC \textit{Planet Earth}. Each thumbnail represents the middle frame of a shot, and the first row is the ground truth segmentation. A change in color underlines a change of story.

Compared to the human annotation, our method identifies the exact change point in four cases, and merges together adjacent ground truth stories in one case. On the other hand, the STG method in~\cite{sidiropoulos11} is able to identify some story changes correctly, but creates short stories with just one shot. The NW method in~\cite{chasanis09} does not show oversegmentation phenomena, but creates unreasonable story changes. Finally, the Siamese approach of~\cite{acm} can actually identify correct story boundaries in some cases, still the segmentation provided by our method looks more consistent with the human annotation.

\begin{figure*}[tb]
    \centering
    \subfigure{
        \includegraphics[width=0.09\textwidth]{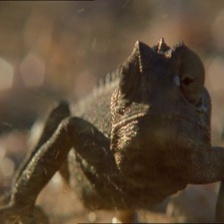}
	}
    \subfigure{
        \includegraphics[width=0.09\textwidth]{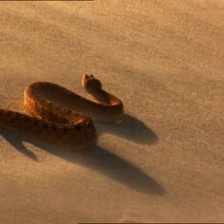}
	}
    \subfigure{
        \includegraphics[width=0.09\textwidth]{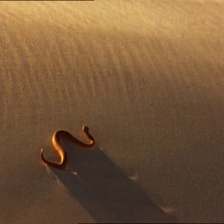}
	}
    \subfigure{
        \includegraphics[width=0.09\textwidth]{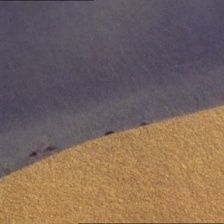}
	}
    \subfigure{
        \includegraphics[width=0.09\textwidth]{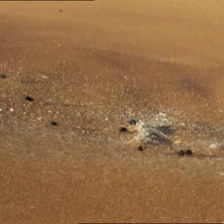}
	}
    \subfigure{
        \includegraphics[width=0.09\textwidth]{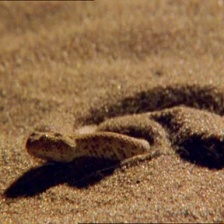}
	}
    \subfigure{
        \includegraphics[width=0.09\textwidth]{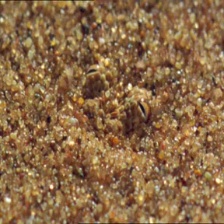}
	}
    \subfigure{
        \includegraphics[width=0.09\textwidth]{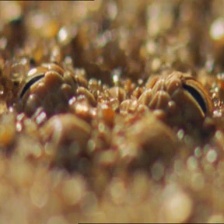}
	} \\
        \subfigure{
        \includegraphics[width=0.09\textwidth]{video_10_shot_057.jpg}
	}
    \subfigure{
        \includegraphics[width=0.09\textwidth]{video_10_shot_059.jpg}
	}
    \subfigure{
        \includegraphics[width=0.09\textwidth]{video_10_shot_058.jpg}
	}
    \subfigure{
        \includegraphics[width=0.09\textwidth]{video_10_shot_062.jpg}
	}
    \subfigure{
        \includegraphics[width=0.09\textwidth]{video_10_shot_064.jpg}
	}
    \subfigure{
        \includegraphics[width=0.09\textwidth]{video_10_shot_061.jpg}
	}
    \subfigure{
        \includegraphics[width=0.09\textwidth]{video_10_shot_060.jpg}
	}
    \subfigure{
        \includegraphics[width=0.09\textwidth]{video_10_shot_063.jpg}
	} \\
    
    \subfigure{
        \includegraphics[width=0.09\textwidth]{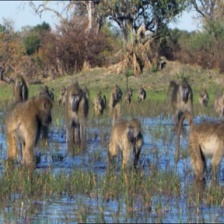}
	}
    \subfigure{
        \includegraphics[width=0.09\textwidth]{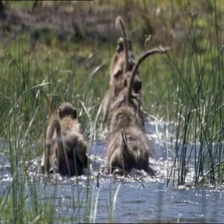}
	}
    \subfigure{
        \includegraphics[width=0.09\textwidth]{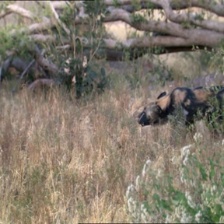}
	}
    \subfigure{
        \includegraphics[width=0.09\textwidth]{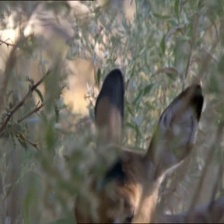}
	}
    \subfigure{
        \includegraphics[width=0.09\textwidth]{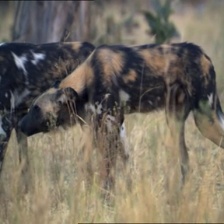}
	}
    \subfigure{
        \includegraphics[width=0.09\textwidth]{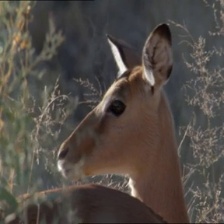}
	}
    \subfigure{
        \includegraphics[width=0.09\textwidth]{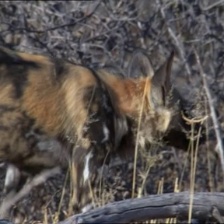}
	}
    \subfigure{
        \includegraphics[width=0.09\textwidth]{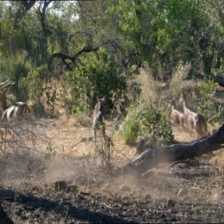}
	} \\    

    \subfigure{
        \includegraphics[width=0.09\textwidth]{video_00_shot_348.jpg}
	}
    \subfigure{
        \includegraphics[width=0.09\textwidth]{video_00_shot_344.jpg}
	}
    \subfigure{
        \includegraphics[width=0.09\textwidth]{video_00_shot_347.jpg}
	}
    \subfigure{
        \includegraphics[width=0.09\textwidth]{video_00_shot_343.jpg}
	}
    \subfigure{
        \includegraphics[width=0.09\textwidth]{video_00_shot_346.jpg}
	}
    \subfigure{
        \includegraphics[width=0.09\textwidth]{video_00_shot_345.jpg}
	}
    \subfigure{
        \includegraphics[width=0.09\textwidth]{video_00_shot_349.jpg}
	}
    \subfigure{
        \includegraphics[width=0.09\textwidth]{video_00_shot_350.jpg}
	} \\    
    
    \caption{Ranking of two sample shot sequences. First and third row report all shots from the two sequences in temporal order, while second and fourth row show the produced aesthetic ranking in descending order (with leftmost thumbnails predicted to be more aesthetically pleasing than rightmost ones). Thumbnails with a centered and clearly visible object are preferred against blurred and low-quality frames (best viewed in color).}
    \label{fig:ranking}
\end{figure*}

\subsection{Evaluation with multiple annotators}
As stated at the beginning of this section, we extended the BBC Planet Earth dataset by collecting four more annotations. This, along with that provided in~\cite{acm}, results in a set of five different annotations, which are used to investigate both the role of subjectivity in story detection, and the capabilities of our embedding to learn a particular annotation style. The choice of this particular dataset is motivated by the fact that in documentaries story boundaries are less objective than in movies and TV-shows, and are also related to changes in topic. Collected annotations differ in terms of granularity (with some annotators putting story boundaries for minor topic or place changes, and others building longer stories) and also in terms of localization (given that sometimes the exact change point is not easy to identify).

We first run the algorithm described in Section~\ref{sec:agreement_dp} to get the the segmentation which maximally agrees with all the given annotations. The resulting segmentation presents a mean $\overline{\text{IoU}}$ with the five annotators of 0.762. This represents an upper-bound for story detection algorithms trained on this set of annotations, given that no segmentation could achieve a better result (ignoring the approximation introduced by our algorithm, which is negligible).

The proposed embedding is then trained and tested on all annotations, as well as on the agreement given by the Dynamic Programming algorithm, always keeping a leave-one-out setup among the eleven videos. Results are reported in Table~\ref{tab:results_subjective}: clearly, higher $\overline{\text{IoU}}$ values are obtained when training and testing on the same annotator, and this suggests that our model was indeed able to capture some features of the segmentation style of an annotator, such as the level of granularity. At the same time, training on the maximum agreement annotation leads, on average, to better $\overline{\text{IoU}}$ scores when testing on the five human annotators.

\subsection{Thumbnail selection evaluation}
On a different note, we conducted a series of experiments regarding the proposed retrieval strategy.
Since aesthetic quality is subjective, three different users were asked to mark all keyframes either as aesthetically relevant or non relevant for the story they belong to. For each shot, the middle frame was selected as keyframe. Annotators were instructed to consider the relevance of the visual content as well as the quality of the keyframe in terms of color, sharpness and blurriness. Each keyframe was then labeled with the number of times it was selected, and a set of $(d_i,d_j)$ training pairs was built according to the given ranking, to train our aesthetic ranking model.

For comparison, an end-to-end deep learning approach (\textit{Ranking CNN}) was also tested. In this case the last layer of a pre-trained VGG-16 network was replaced with just one neuron, and the network was trained to predict the score of each shot, with a Mean Square Error loss. Both the Ranking CNN model and the proposed Hypercolumn-based ranking were trained in a leave-one-out setup, using ten videos for training and one for test from the BBC Planet Earth collection.
 
\begin{table}[tbp]
    \centering
    \caption{Aesthetic ranking: average percent of swapped pairs on the BBC Planet Earth dataset (lower is better).}

    \begin{tabular}{|c|c|c|}
        \hline
        \multirow{2}{*}{\textbf{Episode}} & \multirow{2}{*}{\parbox{2.2cm}{\centering{\textbf{Ranking CNN}}}} & \multirow{2}{*}{\parbox{2.2cm}{\centering{\textbf{Hypercolumns Ranking}}}} \\
        & & \\ \hline
        From Pole to Pole & 8.23  & 4.10 \\ \hline
        Mountains         & 12.08 & 7.94 \\ \hline
        Fresh Water       & 12.36 & 8.11 \\ \hline
        Caves             & 9.98  & 8.76 \\ \hline
        Deserts           & 13.90 & 9.35 \\ \hline
        Ice Worlds        & 6.62  & 4.33 \\ \hline
        Great Plains      & 10.92 & 9.63 \\ \hline
        Jungles           & 12.28 & 7.43 \\ \hline
        Shallow Seas      & 10.91 & 6.22 \\ \hline
        Seasonal Forests  & 9.47  & 4.82 \\ \hline
        Ocean Deep        & 10.73 & 5.75 \\ \hline \hline
        \textbf{Average} & \textbf{10.68} & \textbf{6.95} \\ \hline
    \end{tabular}
    \label{tab:ranking}
\end{table}

Table~\ref{tab:ranking} reports the average percent of swapped pairs: as it can be seen, our ranking strategy is able to overcome the Ranking CNN baseline and features a considerably reduced error percentage. This confirms that low and high level features can be successfully combined together, and that high features alone, such as the ones the Ranking CNN is able to extract from its final layers, are not sufficient. Figure~\ref{fig:ranking} shows the ranking results of two shot sequences: as requested in the annotation, the SVM model preferred thumbnails with good quality and a clearly visible object in the middle. Qualitative results are also available in the demo interface hosted at \url{http://imagelab.ing.unimore.it/neuralstory}, where the reader can test the proposed retrieval system on textual queries.

\section{Conclusion}
This paper presented a new approach for story detection in broadcast videos. Our proposal builds a set of domain specific concept classifiers, and learns an embedding space via a Triplet Deep Network, which considers visual as well as textual concepts extracted from the video corpus.
We showed the effectiveness of our approach compared to different techniques via quantitative experiments, and demonstrated the effectiveness of the proposed features. The subjectivity of the task was also taken into account, by demonstrating that the proposed embedding can adapt to different annotators, and by providing an algorithm to maximize the agreement between a set of annotators. As a potential application of story detection, we also introduced its use in retrieval results presentation, allowing the simultaneous use of semantic and aesthetic criteria. 

\section*{Acknowledgment}
Our work is partially founded by the project ``Citt\`a educante'' (CTN01\_00034\_393801) of the National Technological Cluster on Smart Communities (cofunded by the Italian Ministry of Education, University and Research - MIUR). We acknowledge the CINECA award under the ISCRA initiative, for the availability of high performance computing resources and support.

\ifCLASSOPTIONcaptionsoff
  \newpage
\fi

\bibliographystyle{IEEEtran}
\bibliography{IEEEabrv,2016TMM}

%
%

\end{document}